%% file: main.tex
\documentclass[lettersize,journal]{IEEEtran}
\usepackage{amsmath,amsfonts}
\usepackage{algorithmic}
\usepackage{algorithm}
\usepackage{array}
\usepackage[caption=false,font=normalsize,labelfont=sf,textfont=sf]{subfig}
\usepackage{textcomp}
\usepackage{stfloats}
\usepackage{url}
\usepackage{verbatim}
\usepackage{booktabs}
\usepackage{multirow}
\usepackage{graphicx}
\usepackage{cite}
\usepackage{hyperref}
\usepackage{booktabs}
\usepackage{multirow}
\usepackage{bm}
\usepackage{amsfonts}
\usepackage{amsmath}
\def\R{\mathbb{R}}
\def\S{\mathbb{S}}
\def\vec{\bm}
\def\dif{\mathrm{d}}

\def\vn{\vec{n}}
\def\vv{\vec{v}}
\def\vl{\vec{l}}
\def\vx{\vec{x}}

\def\vr{\vec{r}}
\def\vk{\vec{k}}
\def\vc{\vec{c}}
\def\vf{\vec{f}}

\begin{document}

\title{NeuS-PIR: Learning Relightable Neural Surface using Pre-Integrated Rendering}

\author{Shi Mao,
Chenming Wu$^*$,
Zhelun Shen,
Yifan Wang,
Dayan Wu,
and Liangjun Zhang
\thanks{Shi Mao is with Baidu Research and King Abdullah University of Science and Technology.
  	E-mail: shi.mao@kaust.edu.sa}%
\thanks{
    Chenming Wu, Zhelun Shen and Liangjun Zhang are with Baidu Research.
  	E-mail: \{wuchenming, shenzhelun, liangjunzhang\}@baidu.com
} %
\thanks{
Yifan Wang is with the Paul G. Allen School of Computer Science, University of Washington.
  	E-mail: yifan1@cs.washington.edu.
}
\thanks{
Dayan Wu is with the Institute of Information Engineering, Chinese Academy of Science, Beijing 100084, China (e-mail: wudayan@iie.ac.cn).}
\thanks{C. Wu is the corresponding author.}
\thanks{Manuscript received March 25, 2024.}}

\markboth{}
{Shell \MakeLowercase{\textit{et al.}}: A Sample Article Using IEEEtran.cls for IEEE Journals}

\IEEEpubid{}

\maketitle

\begin{abstract}
This paper presents a method, namely NeuS-PIR, for recovering relightable neural surfaces using pre-integrated rendering from multi-view images or video. 
Unlike methods based on NeRF and discrete meshes, our method utilizes implicit neural surface representation to reconstruct high-quality geometry, which facilitates the factorization of the radiance field into two components: a spatially varying material field and an all-frequency lighting representation. This factorization, jointly optimized using an adapted differentiable pre-integrated rendering framework with material encoding regularization, in turn addresses the ambiguity of geometry reconstruction and leads to better disentanglement and refinement of each scene property.
Additionally, we introduced a method to distil indirect illumination fields from the learned representations, further recovering the complex illumination effect like inter-reflection.
Consequently, our method enables advanced applications such as relighting, which can be seamlessly integrated with modern graphics engines.
Qualitative and quantitative experiments have shown that NeuS-PIR outperforms existing methods across various tasks on both synthetic and real datasets. Source code is available at \href{https://github.com/Sheldonmao/NeuSPIR}{https://github.com/Sheldonmao/NeuSPIR}.
\end{abstract}


\section{Introduction}
\IEEEPARstart{R}{ecovering} 
an object's geometry, material properties, and illumination from multi-view images or video, commonly referred to as \textit{inverse rendering}, is a longstanding challenge in computer vision. These recovered properties are keys to several useful applications such as view synthesis, relighting, and object insertion.
However, inverse rendering is a challenging task due to its fundamentally underconstrained nature. Efforts to address this have involved leveraging additional observations, such as scanned geometry, known lighting conditions, multiple images with varying lighting conditions, or restrictive assumptions of assuming a single material for the object\cite{bi2020deep,bi2020neural,zhang2022iron}.

\begin{figure}[t]
    \includegraphics[width=\linewidth]{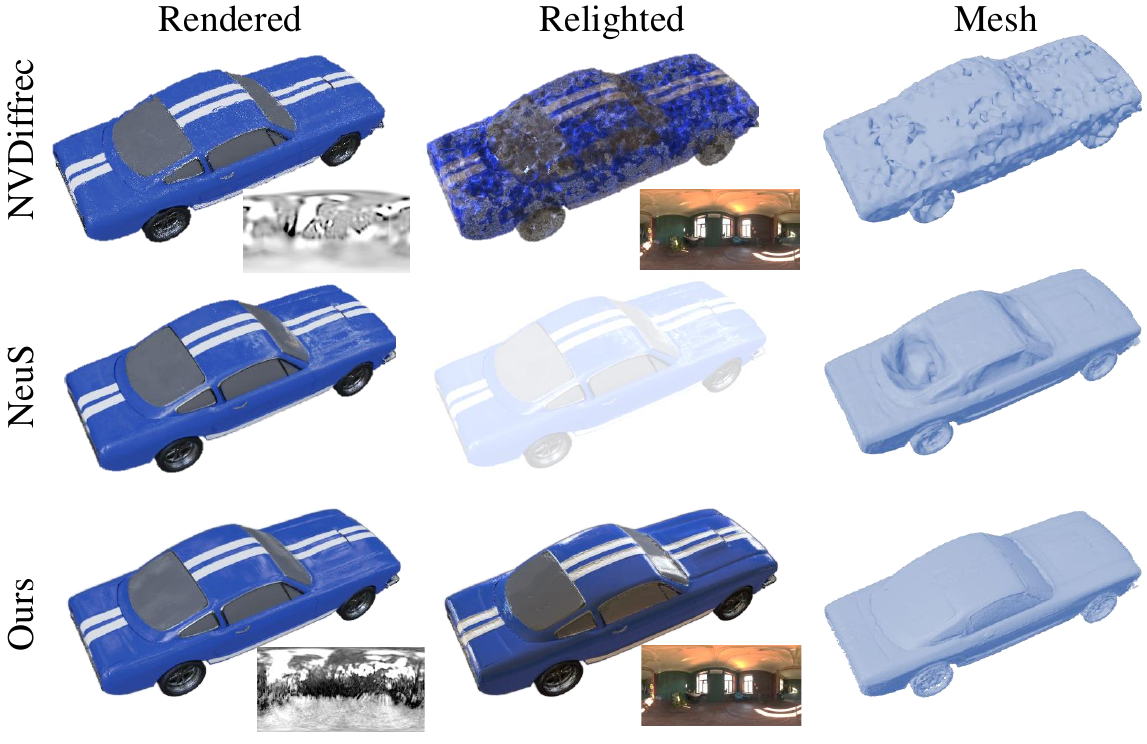}
    \caption{In our proposed method, we achieve simultaneous learning of geometry, material, and illumination within the neural implicit field. The results, displayed in the bottom row, demonstrate significant improvements compared to NVDiffrec~\cite{munkberg2022extracting} showcased in the top row. Our method excels in relighting the image and reconstructing geometry. By building upon NeuS~\cite{wang2021neus}, a popular approach for geometry reconstruction without factorization, our mesh incorporates the advantages of material and illumination learning, resulting in enhanced geometry preservation, particularly in highly reflective areas.}
    \label{fig:teaser}
    \vspace{-10pt}
\end{figure}

The emerging trend of neural representations has demonstrated remarkable capabilities in geometry reconstruction~\cite{wang2021neus} and view synthesis~\cite{mildenhall2020nerf}. In these neural representation-based methods, material properties and illumination are often intertwined. Consequently, they cannot be directly applied to tasks like relighting or material editing. While there have been attempts to decouple neural representations~\cite{zhang2021nerfactor,boss2021neural,srinivasan2021nerv}, major limitations are preventing them from being readily used for relighting. As NeRF-based methods model geometry as a volume density field without sufficient surface constraints, it contains artifacts that may not be noticeable in view synthesis but problematic in extracted high-quality surfaces with reliable normal. 
\textit{Signed Distance Function} (SDF) is introduced into volume rendering to supervise the geometry depiction explicitly and has achieved improved results~\cite{wang2021neus,oechsle2021unisurf,yariv2021volume}. However, these methods focus more on geometry reconstruction rather than inverse rendering. 
Other methods like NVDiffrec~\cite{munkberg2022extracting} adopt hybrid approach that uses implicit
SDF field and explicit mesh and proposes an efficient differential rendering pipeline for reconstruction, but such geometry representation might lead to inaccurate topology due to predefined SDF-grid.
Overall, existing inverse rendering approaches~\cite{boss2021nerd,zhang2021nerfactor,munkberg2022extracting} mostly focus on decomposing material and illumination while geometry quality is not prioritized. 

In this work, we demonstrate that it is possible to recover fine-grained geometry details, convincing material properties, and realistic illuminations at the same time, as shown in Fig.~\ref{fig:teaser}. Our proposed method, which we call NeuS-PIR, takes advantage of an implicit geometry representation of NeuS, and a novel pre-integrated rendering (PIR) network for the joint optimization.
The adopted NeuS~\cite{wang2021neus} geometry representation facilitates the reconstruction of smooth surfaces and well-defined surface normal, which makes joint optimization both feasible and stable. NVDiffrec's discretized mesh-based geometry representations heavily rely on a good initialization from the first stage and often struggle to reconstruct smooth surfaces. 
For illumination integration, NeuS-PIR advances the pre-integrated rendering technique, allowing it to learn all frequency illumination through a differentiable high-frequency environment map (\emph{i.e.} a $6 \times 512 \times 512$ cube map), which leads to better reconstruction quality for shiny surfaces. Existing methods typically resort to low-frequency illumination to reduce the computational costs, \emph{e.g.} pre-integrated rendering using low-dimensional envmap embedding in Neural-PIL and Sphere Gaussian in PhySG, which sacrifices the quality of the recovered illumination. 
A radiance field is proposed to guide the learning process, accompanied by customized regularization on geometry, lighting, and materials to ensure stable training. Additionally, an indirect illumination field can be distilled from the learned representations to further address the complex lighting. All representations are jointly optimized through differentiable pre-integrated rendering.  
In summary, our  main technical contributions are:
\begin{itemize}
  \item a framework that leverages a neural implicit surface and pre-integrated rendering to factorize the scene into geometry, material, and illumination, leading to less degraded geometry and better disentanglement, thereby supporting relighting,
  \item a joint optimizing scheme using pre-computed environment map with material encoding regularization that encourages sparsity and consistency, which enables all-frequency illumination recovery with improved generalization ability compared to data-driven latent-space methods~\cite{boss2021neural,zhang2021nerfactor}, yet can be optimized efficiently, and
  \item a modular distillation method to distill indirect illumination fields from learned representations by jointly optimizing it with the direct illumination, which further addresses the complex lighting effects like inter-reflection.
\end{itemize}

\section{Related Work}
\subsection{Multi-view Reconstruction}

\vspace{2pt} \noindent \textbf{Explicit Reconstruction} uses 
depth sensors or multi-view stereo methods~\cite{yao2018mvsnet,shen2022pcw,9797764, zhang2021farther, 9504486, 9360611}, which estimates the depth maps
by matching feature points across different views, to reconstruct images using triangle mesh~\cite{liu2019soft}, tetrahedral mesh~\cite{munkberg2022extracting}, voxel~\cite{liao2018deep}, hierarchical octree~\cite{hane2017hierarchical}, atlas surface~\cite{groueix2018papier}, or explicit/implicit hybrid~\cite{shen2021deep}. One significant advantage of explicit representation is the reconstructed models are compatible with downstream rendering on industrial engines. 
However, optimizing explicit representation is sensitive to hyperparameters and initialization, sometimes resulting in reconstruction failures due to topological inconsistency. In contrast, neural implicit fields are more stable.

\vspace{2pt} \noindent \textbf{Neural Implicit Field Modeling} uses neural networks to represent spatial numerical fields for modeling 3D objects or scenes. Neural Radiance Field (NeRF)~\cite{mildenhall2020nerf} and its variants (\emph{e.g.}, MipNeRF~\cite{barron2021mip}) use coordinate-based MLP to spatially encode the volumetric radiance space, and geometry and color of an arbitrary point inside this space can be queried by MLP, allowing for rendering by casting rays and integrating all the queried values. While NeRF-like methods can synthesize high-quality novel views, the intrinsic geometry is not explicitly optimized. Recent advances in neural surface reconstruction carefully design the optimization framework that can optimize the geometry using photometric loss. For example, UniSurf~\cite{oechsle2021unisurf} proposes to sharpen the sampling distribution to align the volumetric field to the surface.
VolSDF~\cite{yariv2021volume} converts the density function in NeRF to a learnable SDF transformation and samples points along the casting rays according to the error bound of opacity.
NeuS~\cite{wang2021neus} and its follow-up~\cite{wang2022neus2,long2022sparseneus} provide an unbiased and occlusion-aware solution to convert the density function in NeRF to SDF. 
Although these methods can reconstruct plausible geometry and render high-quality novel views, the material, and illumination are baked into the model itself, resulting in unsatisfactory rendered results when background scenes are different. We extend NeuS with the ability to factorize material and illumination for downstream applications like relighting and object insertion.

\input{tables/tab_methods}
\subsection{Material and Illumination Estimation}

Estimating material and illumination for reconstructed objects is a challenging task. Previous methods require known lighting conditions~\cite{zeng2023nrhints} or estimating light source positions using~\cite{wang2022neural,8936483,9761930} for inverse rendering.
\cite{bi2020deep,bi2020neural} use differentiable volume ray marching framework to supervise reconstructing a neural reflectance volume and reflectance field, respectively. 
Deferred Neural Lighting~\cite{gao2020deferred} applies deferred rendering using proxy geometry and neural texture followed by neural rendering to enable free-viewpoint relighting. 
Another mesh-based method~\cite{luan2021unified} jointly optimizes mesh and SVBRDF by a differentiable renderer specialized for collocated configurations.
IRON~\cite{zhang2022iron} proposes a two-stage method that first uses a signed distance field to recover geometry and then optimize material. 
These methods require photometric images, leading to a more involved data-capturing process, while our method jointly optimizes geometry, material, and illumination with only images as input.

Recent efforts on inverse rendering enable us to estimate material and illumination from relaxed multi-view settings. 
NeRFactor~\cite{zhang2021nerfactor} uses a set of MLPs to describe light source visibility, normal maps, surface albedo, and the material property on the surface point, on top of a pre-trained NeRF model. 
PhySG~\cite{zhang2021physg} sets a precondition that the scene is under a fixed illumination.
NeRD~\cite{boss2021nerd} extends the fixed illumination condition to both fixed or varying illumination.
Neural-PIL~\cite{boss2021neural} proposes a neural pre-integrated lighting method to replace the spherical Gaussians, which enables estimating high-frequency lighting details.
Ref-NeRF~\cite{verbin2022ref} 
replaced MipNeRF~\cite{barron2021mip}'s parametrization of view-dependent outgoing radiance with a reflected radiance to model environment light and surface roughness. 
NeILF~\cite{yao2022neilf} proposes to use a fully 5D light field to model illuminations of any static scene, where occlusions and indirect lights are handled naturally. NeILF++~\cite{zhang2023neilf++} eliminates the requirement of reconstructed geometry as an input.
Instead of only using implicit fields to describe the scenes, NVDiffrec~\cite{munkberg2022extracting} adopts a hybrid approach that uses implicit SDF field and explicit mesh and proposes an efficient differential rendering pipeline for reconstruction. NVDiffrecmc~\cite{hasselgren2022nvdiffrecmc} adopts the same geometry and material representations as NVDiffrec but incorporates ray tracing and Monte Carlo integration for more realistic shading. 
Lin et al.~\cite{9915626} propose to use physically-based rendering to recover texture for a given mesh. TensoIR~\cite{jin2023tensoir} opts to use tensor factorization and neural fields rather than using MLP-based neural fields. ENVIDR~\cite{liang2023envidr} proposes an implicit differentiable render for inverse rendering, achieving impressive performance, but difficult to use in modern graphics pipelines, for example, relighting with an off-the-shelf HDR envmap without training. \cite{10210281} investigates the scene decomposition under dynamic point light sources.
There are a few concurrent works: NeRO~\cite{liu2023nero} uses a similar geometry representation and learning strategy. Their focus is to recover reflective objects, while ours is designed for the inverse rendering of general objects; NeuralPBIR~\cite{sun2023neural} proposes a two-stage pipeline that firstly reconstructs an imperfect geometry and then applies physics-based inverse rendering for higher quality factorization, while they dismiss learning metallic parameters using a simplified BRDF rendering, thus hard to apply to the objects with highly reflective surfaces, such as the toaster Fig.~\ref{fig:shiny_visualize} shown in our work. Recent advances in 3D Gaussian Splatting (3DGS)~\cite{kerbl20233d} bring new representation and factorization in inverse rendering (\emph{e.g.}, \cite{jiang2023gaussianshader,gao2023relightable,shi2023gir,liang2023gs}). Differently, Gaussian is an under-defined explicit geometry, while the representation of mesh is more suitable for geometric manipulation, such as deformation, etc.

\begin{figure*}[t]
    \includegraphics[width=\textwidth]{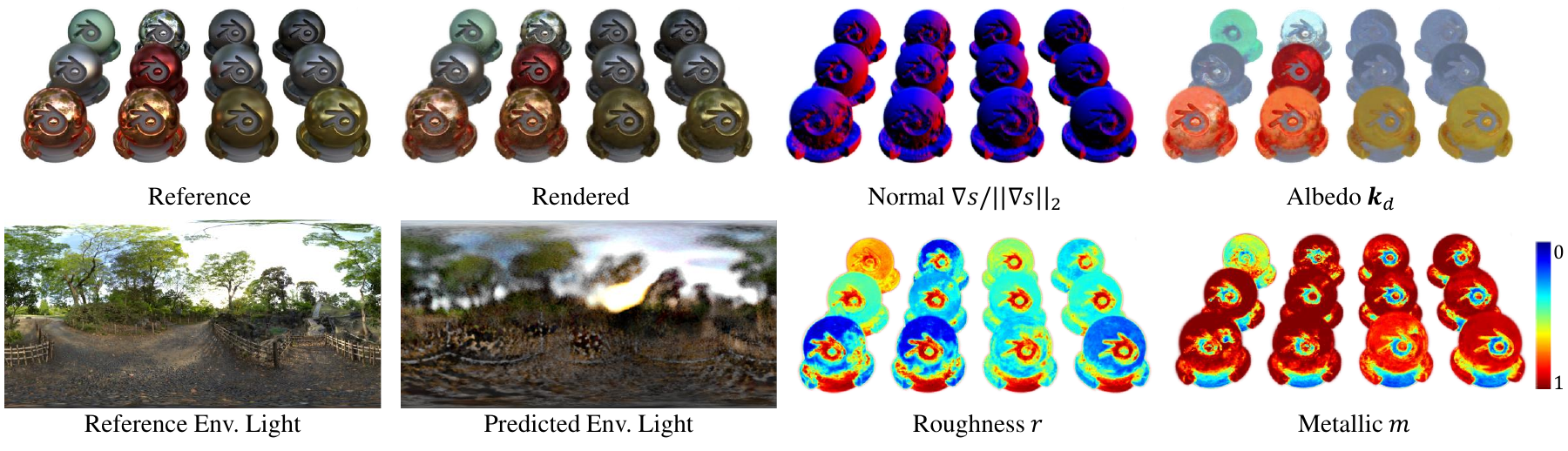}
    \caption{In our approach, we decompose a scene into three components: geometry, material, and illumination. To visualize the results, we present the reference and predicted environment illuminations as latitude-longitude converted environment cubemaps. The roughness and metallic properties are visualized using a jet color map, which represents values ranging from 0 to 1.}
    \label{fig:method-overview}
\end{figure*}

\section{Methodology}
Given a collection of multi-view images from images or monocular video and their corresponding camera poses, we aim to reconstruct the object in camera views and its surrounding environment illumination. To be concrete, our factorization method learns an implicit representation of the object-related geometry and material properties and the object-irrelevant illumination. 
As shown in Fig.~\ref{fig:method-overview}, for each point $\vx \in \R^3$ as an input, our model outputs its \textit{signed distance field} (SDF) value $s \in \R$, diffuse albedo $\vec{k_d} \in [0,1]^3$, roughness $r \in [0,1]$ and metallic $m \in [0,1]$, as well as the illumination represented by an environmental cubemap $I \in \R^{H,W,6}$.  
After the training is finalized, our method enables us to render reconstructed objects under different environment maps representing the illumination. This is referred to as \textit{relighting} and can be achieved by jointly estimated or externally defined environment maps. The rendering process follows the volume rendering principle. In addition, high-quality surface mesh with material properties can be easily obtained from the learning representation using an off-the-shelf mesh approximation algorithm (\emph{i.e.}, marching cube or its variants), and enables downstream applications.

\subsection{Learning Geometry}

Our method adopts Neural Implicit Surfaces (NeuS)~\cite{wang2021neus} as the representation of an object's geometry, considering its ability to reconstruct high-quality surfaces as the zero-level set of implicit SDF representation. NeuS uses multi-layer perceptron (MLP) to learn both the SDF function $f_{sdf}:\vx \mapsto s$ that maps a 3D position $\vx \in \R^3$ to an SDF value $s \in \R$, and appearance mapping $f_{color}:(\vx,\vec{v}) \mapsto \vec{c}$ that maps a 3D position $\vx \in \R^3$ and viewing direction $\vec{v} \in \S^2$ to its corresponding RGB radiance $L \in [0,1]^3$.

NeuS renders an image by accumulating the radiance along the rays cast by pixels, following the standard volume rendering scheme. Specifically, given a pixel ray parameterized as $\left\{ \vx(t) = \vec{o} - t\vec{v} | t \geq 0 \right\}$, where $\vec{o} \in \R^3$ %
represents its corresponding camera origin, and $\vec{v} \in \S^2$ is its normalized direction pointing towards the camera center, the accumulated color for this pixel $\vec{c}_p$ can be computed as a weighted sum of colors along the ray: 

\begin{equation}
    \vec{c}_p(\vec{o},\vec{v})=\int_{t_n}^{t_f} w(t) L\left(\vx(t), \vec{v} \right) \dif t,
\label{eq:neus-render}
\end{equation}
where $w(t)$ is a non-negative weight function, and we integrate it from near plane $t_n$ to far plane $t_f$ of the camera model. By enforcing the unbiased and occlusion-aware requirements, NeuS derives the weight function from SDF as:
\begin{equation}
    w(t) =  \exp\left( -\int_0^t \rho(u) \dif u \right) \rho(t),
\label{eq:neus-weight}
\end{equation}
where $\rho(t)=\max \left(\frac{\frac{- \dif \Phi_\tau}{ \dif t}\left(s(t)\right)}{\Phi_s\left(s(t)\right)},0\right)$ is refereed to as opaque density, and $\Phi_\tau(s)=(1+e^{- \tau s})^{-1}$ is the Sigmoid function scaled by a factor $\tau$. The learned factor $\tau$ is inversely proportional to the standard deviation of density distribution near the zero level across the SDF. During training, $1/\tau$ is expected to converge to zero as the zero-valued isosurfaces of the SDF gradually approach solid surfaces.

To enable relighting and material factorization, we employ a dual-branch approach to model the outgoing radiance along the ray cast from a pixel. In the first branch, the radiance MLP directly generates the outgoing radiance based on the position, viewing direction, and surface normal. Meanwhile, the second branch, which is material-aware, considers material and illumination properties and employs a split-sum approximation method outlined in Section~\ref{sec:method-material} to render the outgoing radiance. Both branches utilize the same SDF module to maintain consistency in geometry throughout the learning process.

\subsection{Learning Material and Illumination}
\label{sec:method-material}

We adopt image-based lighting as our lighting model to decompose the radiance field into geometry, material, and lighting components and approximate the rendering equation with pre-integrated rendering. Following~\cite{karis2013ue4}, the specular term in rendering equation can be approximated by \textit{split sum} approximation:
\begin{equation}
    \int_{\Omega} L_i(\vl)f_s(\vl,\vv)(\vl\cdot \vn) \dif \vl = I(\vr;r) \int_{\Omega}f_s(\vl,\vv)(\vl\cdot \vn) \dif \vl,
\label{eq:PIR-splitsum}
\end{equation}
where $L_i(\vl)$ represents the incident radiance from light direction $\vl$, $f_s(\vl,\vv;r,m)$ denotes Cook-Torrance \cite{cook1982reflectance} microfacet specular BRDF, parameterized by roughness $r$ and metallic $m$, and $\vn$ is the surface normal vector. The first term in split sum approximation $I(\vr;r)$ involves an importance sampling of incident light radiance mediated by the surface roughness $r$. At a minor cost of losing lengthy reflection at grazing angles, this term 
can be pre-integrated from the environment map and queried from the reflection direction $\vr = 2(\vv \cdot \vn)\vn - \vv$ as:
\begin{equation}
        I(\vr;r) = \int_{\Omega}L_i(\vl)D(\vl,\vr; r)(\vl\cdot \vr) \dif \vl,
\label{eq:PIR-1}
\end{equation}
where $D(\vl,\vv; r)$ represents the normal distribution of GGX~\cite{walter2007GGXmicrofacet}, which takes into account the percentage of microfacets that reflect light towards the viewer and is defined by the roughness $r$. The pre-integrated illumination is represented as the mipmap levels of a learnable environment cubemap.

The second term of the equation is irrelevant to illumination and is equivalent to integrating specular BRDF $f_s$ in a constant brightness environment. Using Schlick's equation, the specular reflectance at normal incidence $F_0$ can be factorized out. Therefore, the second term can be rewritten as $F_0$ modulated by its scale and bias, which are only related to the material's roughness and the cosine between the viewing angle and the surface's normal vector $(\vv \cdot \vn)$.
\begin{equation}
    \int_{\Omega}f_s(\vl,\vv)(\vl\cdot \vn) \dif \vl = F_0 S \left((\vv \cdot \vn) , r \right) + B \left((\vv \cdot \vn) , r \right),
\label{eq:PIR-2}
\end{equation}
both the scale $S$ and bias $B$ can be pre-calculated and stored as 2D look-up table (LUT) for efficient inference~\cite{karis2013ue4}. We adopt the convention to set $F_0$ as an interpolation from $0.04$ (non-metallic material's specular reflectance) to diffuse color $k_d$ (metallic material's specular reflectance) using the material's metallic value $m$. 
\begin{equation}
    F_0 = 0.04\times(1-m) + m \vk_d.
\label{eq:PIR-F0}
\end{equation}

Finally, the direct outgoing radiance $L_{dir}(\vv)$ observed from the viewing direction $v$ is a blend of diffuse and specular terms formulated as follows.
\begin{equation}
    L_{dir}(\vv)  = \vk_d I_d + I_s \left( F_0 S \left((\vv \cdot \vn) , r \right) + B \left((\vv \cdot \vn) , r \right) \right), \\
\label{eq:PIR-render}
\end{equation}
where $I_d = I(\vn;1)$ denotes the diffuse irradiance and $I_s = I(\vr;r)$ denotes specular irradiance. For more details of pre-integrated rendering, readers can refer to the presentation by UE4 \cite{karis2013ue4}.

\subsection{Network Architecture}
\begin{figure}[!t]
    \includegraphics[width=\linewidth]{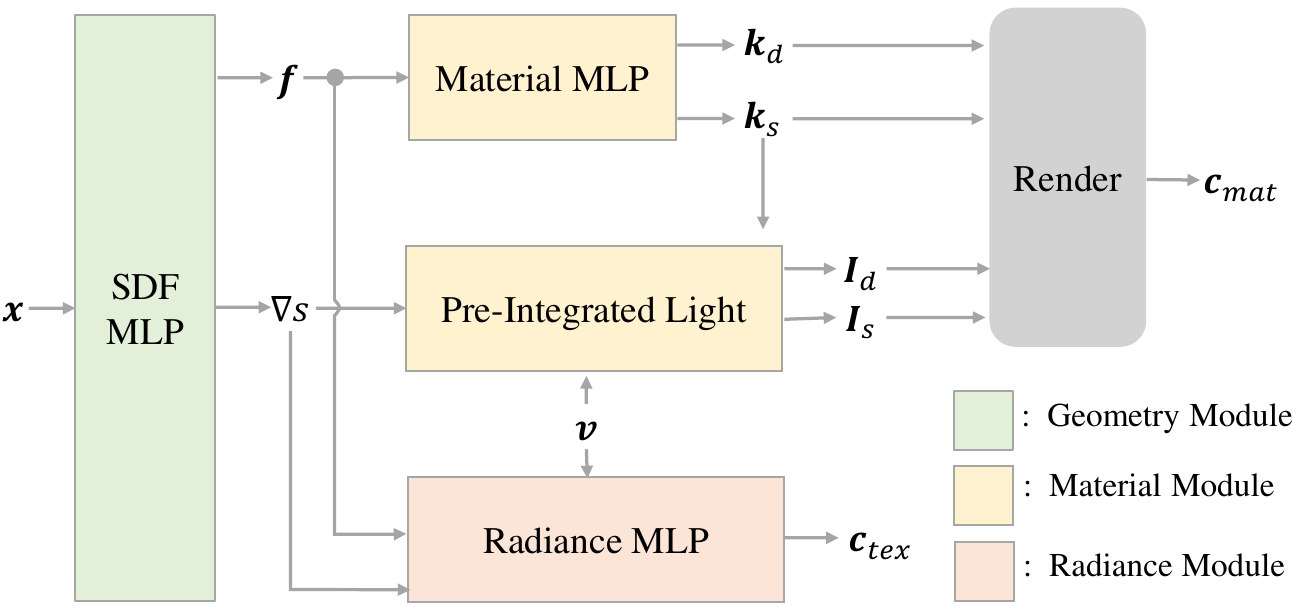}
    \caption{The network architecture of our proposed method consists of multiple components. The SDF MLP is responsible for learning the geometry, while the radiance MLP focuses on capturing the radiance field at a coarse level. Through additional decompositions, we employ the Material MLP to factorize the material properties, and the Pre-Integrated Light module to handle illumination aspects. Together, these components contribute to the overall functionality and capabilities of our method.}
    \label{fig:method-net-arc}
\end{figure}

Fig.~\ref{fig:method-net-arc} illustrates the architecture of our proposed method. The SDF MLP learns the scene's geometry, while the Radiance MLP learns the radiance field of the scene at a coarse level, given the geometry features from SDF MLP and viewing directions. Following the pre-integrated rendering principle, further decompositions are applied to factor the radiance into material and lighting factors.

\textbf{Geometry Module.} The SDF MLP takes 3D positions $\vx$ as input and produces the feature $\vec{f}(\vx)$ as output. The first channel of the output feature represents its SDF value $s(\vx)$, and its gradient $\nabla s(\vx)$ is calculated analytically. Position encoding uses trainable multi-resolution grids that can be efficiently supported by hashtables~\cite{muller2022instant}. The feature is learned through an MLP.

\textbf{Radiance Module.} To initiate the training of the scene's geometry, we begin by determining the view-dependent color for each point using MLP. The radiance MLP takes into account the viewing direction ($\vv$), positional feature ($\vec{f}(\vx)$), and the surface's unit normal vector ($\nabla s(\vx) / ||\nabla s(\vx)||_2$) as input.
The viewing direction is encoded using sphere harmonics up to the $4^{th}$ level. Finally, the output color $\vec{c}_{tex}$ is integrated using Eq.~\ref{eq:neus-render}

\textbf{Material Module.} The material module decomposes view-dependent outgoing radiance as incident light modulated by surface material. The Material MLP, \emph{i.e.} $Mat(\vec{f})$ , takes SDF features $\vec{f}(\vx)$ as input and generates surface diffuse albedo $\vk_d \in [0,1]^3$, metallic $m \in [0,1]$ and roughness $r \in [0,1]$. Following Munkberg et al.~\cite{munkberg2022extracting}, Material MLP additionally produces a general occlusion term $o \in [0,1]$ that takes into account the effect of indirect illumination and shadowing by modulating the computed outgoing radiance by $1-o$. Notice the occlusion term can be refined and distilled from the learned representations in the indirect illumination distillation process detailed in Sec.~\ref{sec:indirect_illum}. We refer to a material's specular and occlusion properties as $\vk_s=\left\{o,r,m\right\}$. Both $\vk_d$ and $\vk_s$ are learned through MLP layers and activated by the sigmoid function to limit their value range to $[0,1]$.

To render the outgoing radiance using pre-integrated illumination, we follow the approach of Munkberg et~al.~\cite{munkberg2022extracting} to use high-resolution cubemap as trainable parameters and pre-integrate $I(\vr;r)$ for discrete roughness levels as its mipmaps. To obtain a specific roughness $r$, we query $I(\vr;r)$ using mipmap interpolation. The view-dependent radiance is then rendered using ~Eq.~\ref{eq:PIR-render} and transformed to the S-RGB space via gamma correction. Finally, we volume render the pixel color $\vc_{mat}$ using ~Eq.~\ref{eq:neus-render}.

\subsection{Loss and Regularization}
We employ Mean Square Error (MSE), L1 loss, and binary cross-entropy loss as supervision measures for the masked rendered images. We also include additional regularization terms for SDF, material, and light. Specifically, image color loss $\mathcal{L}_{\hat{\vc}} = \lambda_{c1}||\hat{\vc} - \vc||_1 +  \lambda_{c2}||\hat{\vc} - \vc||_2$, and $\hat{\vc}$ represents volume rendered pixel color from either material module $\vc_{mat}$ or texture module $\vc_{tex}$. $\mathcal{L}_{mask} = \lambda_{mask} {\rm BCE}(mask,opa)$ is the binary cross entropy of the image mask and the accumulated opaque density along pixel rays.

To regularize the SDF field, we use Eikonal and sparsity terms, \emph{i.e.} $\mathcal{L}_{sdf} = \lambda_{se} ||\nabla s - 1||_2^2 + \lambda_{ss} \exp(-\lambda_{sa} |s|_1)$. The first Eikonal term encourages the gradient of the SDF field to have a unit length, and the second term encourages the zero-valued level crossing of the SDF field to be sparse. These regularization terms encourage learning smooth and sparse solid surfaces.

Material estimation is hard due to limited observation of each surface point. However, this problem can be mitigated by introducing the prior that objects are usually made of a limited number of distinct materials. Following this principle, we regularize the material representation in feature space, promoting smoothness and sparsity, and in image space, promoting local consistency. 

Specifically, in feature space, we adopt material feature loss similar to~\cite{zhang2022invrender} in regularizing material-related latent feature $f$ to be sparse, and the mapping of Material MLP is smooth with regard to small changes in the latent feature space:

\begin{equation}
\begin{aligned}
    \mathcal{L}_{mat}^F = & \lambda_{mf1}\sum_{i=2}^F D_{KL}({\rm Bern}(0.05) || p(f_i)) \\
    &+ \lambda_{mf2}||{\rm Mat}(\vf)-{\rm Mat}(\vf+\Delta \vf)||_1,
\end{aligned}
\end{equation}
where $p(\vf_i)$ is calculated as the mean value of i-th channel of the positional feature $\vf$, which represents the probability of non-zero values. The sparsity loss minimizes its KL-divergence with a target Bernoulli distribution with probability $0.05$, encouraging zero-values for $F-1$ feature channels (excluding the first channel representing the SDF value). The smoothness loss encourages similar latent features (differentiated by a small $\Delta \vf \sim \mathcal{N}(0,\epsilon)$) to be mapped to similar material parameters through Material MLP.

For image space regularization, we sample half of our rays using a patch-based method from images to encourage local similarity in roughness and metallic properties. Additionally, we regularize the amplitude of the occlusion. 
\begin{equation}
    \begin{aligned}
        \mathcal{L}_{mat}^{I} &= \sum_{i=1}^P \left(\lambda_{mid}\delta_i(\vk_d)+ \lambda_{mir}\delta_i(r)+\lambda_{mim}\delta_i(m) \right)  
        \\&+ \lambda_{mio}||o||_2^2,
    \end{aligned}
\end{equation}
where $\delta_i$ calculates the standard deviation of i-th image patch. We calculate the full material regularization as: $\mathcal{L}_{mat} = \mathcal{L}_{mat}^{F} + \mathcal{L}_{mat}^{I}$

To regularize the environment cubemap, we apply a white environment prior and regularize it using its mean absolute error (MAE). Specifically, we use the loss function $\mathcal{L}_{light} = \lambda_l {\rm MAE}(I_{base})$, where $I_{base}$ is the learned environment cubemap at the 0-th mipmap level.
In a nutshell, the total loss function we use is the sum of the above losses:

\begin{equation}
    \mathcal{L} =  \mathcal{L}_{\hat{\vc}_{mat}} +\mathcal{L}_{\hat{\vc}_{tex}} + \mathcal{L}_{sdf}  + \mathcal{L}_{mask} +   \mathcal{L}_{mat} + \mathcal{L}_{light}.
\end{equation}

\begin{figure}
    \centering
    \includegraphics[width=0.9\linewidth]{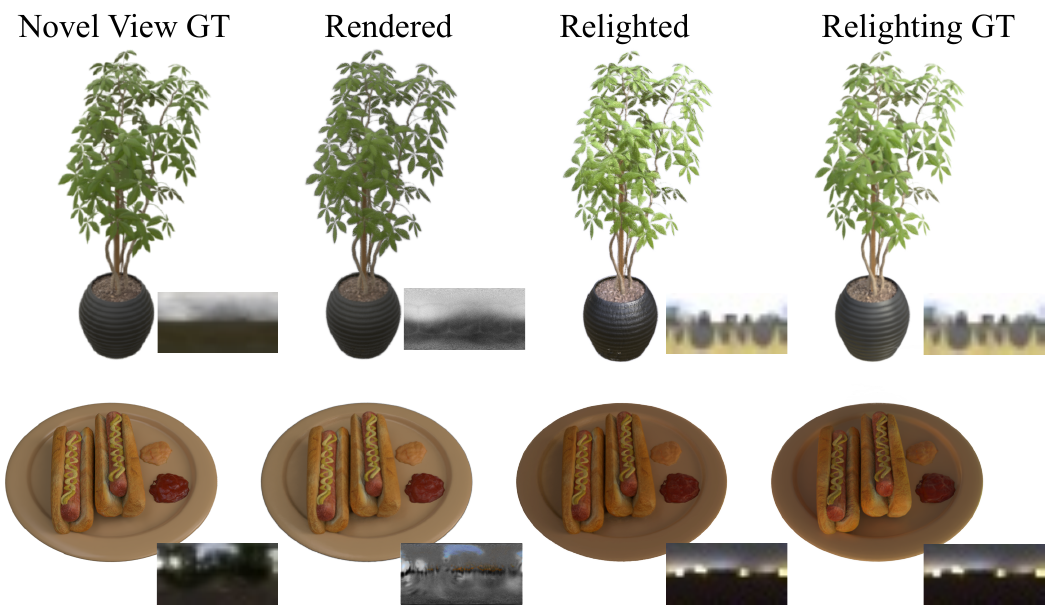}
    \caption{Novel view synthesis and relighting results produced by our proposed method.}
    \label{fig:exp1-relight-viewgen}
\end{figure}

\begin{figure}
    \centering
    \includegraphics[width=\linewidth]{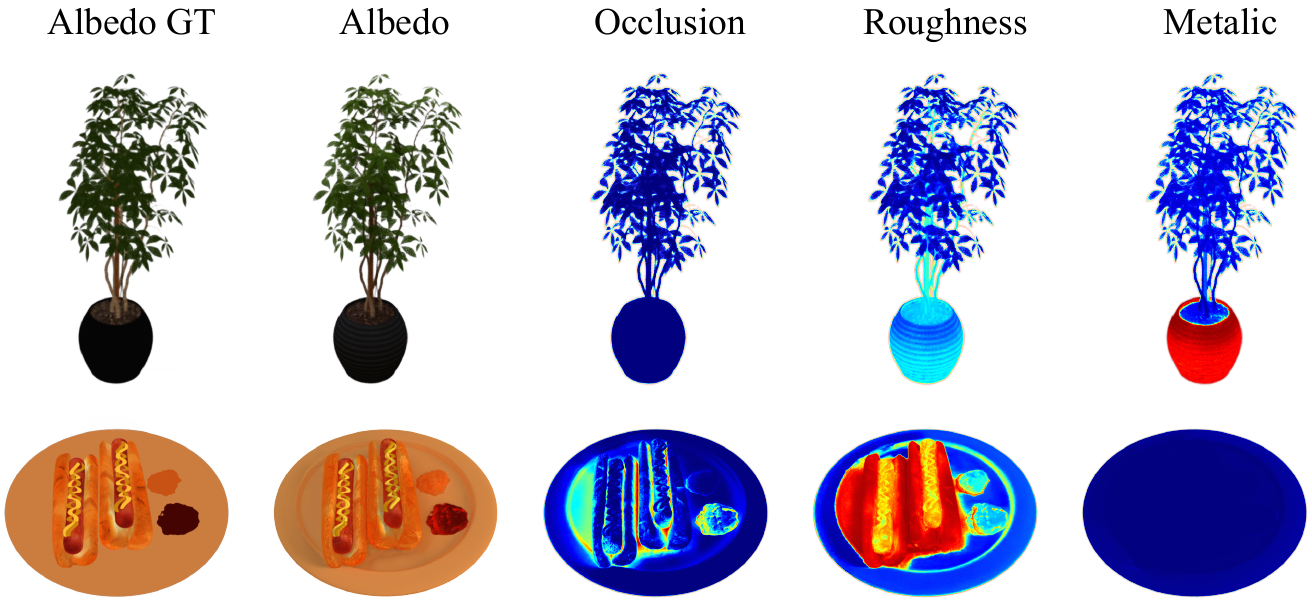}
    \caption{An example of material factorization by our proposed method.}
    \label{fig:exp1-material_factorization}
\end{figure}

\begin{figure*}
    \centering
    \includegraphics[width=\linewidth]{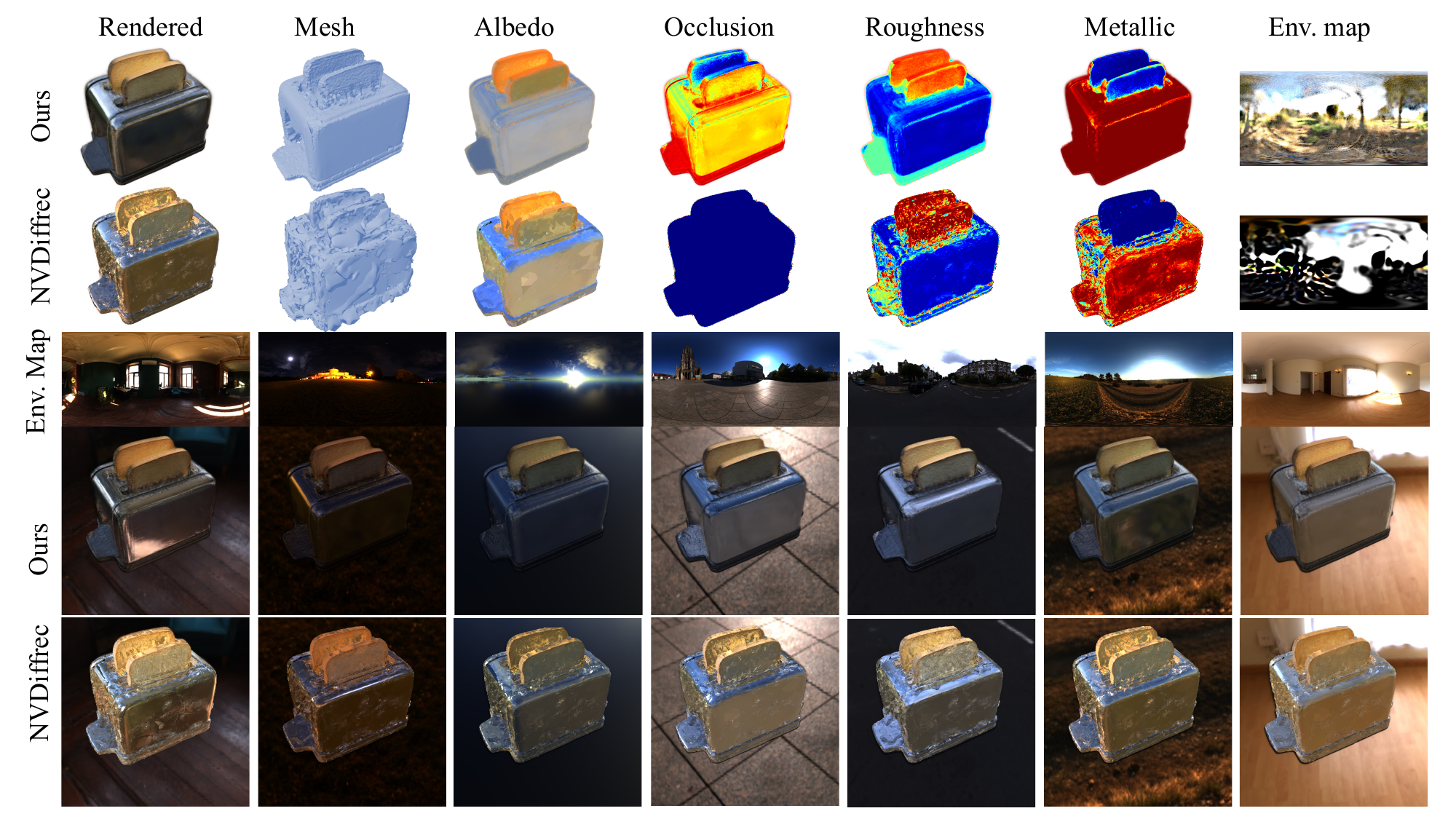}
    \caption{Material factorization and relighting on Ref-NeRF's Shiny scene. From top to bottom: material factorization results by our method; material factorization results by NVDiffrec; high-frequency environment maps used for relighting; our relit results under the given corresponding lights; relit results under the given corresponding lights by NVDiffrec.}
    \label{fig:shiny_visualize}
\end{figure*}

\subsection{Indirect Illumination}
\label{sec:indirect_illum}

In order to tackle the intricate challenges posed by indirect lighting, we introduce an additional step of distilling the indirect illumination field from the accompanying radiance field. This distilled indirect illumination is then incorporated into the rendering process when direct reflections are geometrically obstructed. This post-processing scheme of indirect illumination distillation occurs after the initial training. Similar to the approach in InvRender\cite{zhang2022invrender}, we parameterize the incident indirect illumination using $M=12$ Spherical Gaussians (SGs) for spatial positions:

\begin{equation}
    L_{i}^{ind}(\vec{\omega},\vx) = \sum_{k=1}^M G_{\vx}(\vec{\omega};\vec{\xi}_k,\lambda_k,\vec{\mu}_k),     
\end{equation}
where $\vec{\omega} \in \S^2$ is the queried incident direction at position $\vx$. $\vec{\xi}_k \in \S^2$, $\lambda_k \in \R_+$, and $\vec{\mu}_k \in \R^3$ are the axis, sharpness and amplitude parameters for the Gaussian lobe respectively. The parameters of SGs are modeled using MLP, which takes only position $\vx$ as input. Similarly, a specular occlusion term $o_{s}(\vec{\omega},\vx)$ is modeled to query whether a direction $\vec{\omega}$ is blocked geometrically effectively. The specular occlusion is hemispherically integrated to approximate the diffuse occlusion term $o_d(\vx)$. Following PhySG\cite{zhang2021physg}, the rendering equation can be integrated effectively by approximating GGX normal distribution using SG as well. The composed out-going radiance is computed as a combination of direct and indirect components, each weighted by the specular and diffuse occlusion differently:

\begin{align}
    &L = L_{dir} + L_{ind}, \label{eq:indirect_compose} \\
    &L_{dir} = (1-o_s(r))L_{dir}^{s} + (1-o_d) L_{dir}^{d}, \label{eq:indirect_compose_dir} \\
    &L_{ind} = o_s(r)    L_{ind}^{s} + o_d L_{ind}^{d},  \label{eq:indirect_compose_indir}
\end{align}
where $L_{dir}$ and $L_{ind}$ are direct and indirect out-going radiance, and the superscript $s$ and $d$ stand for the specular and diffuse components of each out-going radiance, respectively. The direct and indirect out-going radiances are weighted by occlusion terms in a complementary manner: When the object's geometry obstructs the incoming light, it is illuminated by indirect illumination. Conversely, when there is no obstruction, the object can be directly illuminated by the environment illumination. For the specular component, we simplify the integration of occluded incident light by only considering the occlusion on the reflection direction $\vr$. We assume that the outgoing specular radiance is primarily influenced by mirror reflection, thus, multiply the occlusion with the outgoing radiance. Likewise, the diffuse occlusion effect is simplified by multiplying the integrated diffuse occlusion with the outgoing radiance, assuming that the incident lights are uniformly occluded. Despite this simplification, the experiments demonstrate reasonably good performance.

To learn the indirect illumination field, we begin by performing hemisphere sampling on the surface point $\vx$ to obtain secondary ray directions $\vec{\omega}_s$. These directions are used in a sphere tracing algorithm to identify intersections with other surfaces, leading to a new surface point $\vx'$. Finally, the Radiance MLP is employed to extract radiance along the opposite sampling direction, which serves as supervision for the corresponding indirect illumination:

\begin{equation}
      \mathcal{L}_{ind} = \lambda_{ind} \sum_{s \in S} || L_{ind}(\vec{\omega}_s,\vx) - L_{tex}(-\vec{\omega}_s,\vx') ||_2^2.
\end{equation}

In a similar manner, the supervision of the specular occlusion term $o_s$ is determined based on whether the sphere tracing algorithm identifies an intersection along the sampled direction $\vec{\omega}_s$, referred to as $o_s'(\vec{\omega}_s,\vx)$. This supervision is achieved through the use of Mean Squared Error (MSE) loss.

\begin{equation}
    \mathcal{L}_{occ} = \lambda_{occ} \sum_{s \in S} || o_s(\vec{\omega}_s,\vx) - o_s'(\vec{\omega}_s,\vx) ||_2^2.
\end{equation}

\section{Experiments}
\subsection{Baselines and Metrics}

Our work is closely related to two methods, namely NVDiffrec~\cite{munkberg2022extracting} and Neural-PIL~\cite{boss2021neural}. Both methods utilize pre-integrated illumination for lighting modeling but differ in their approaches to geometry modeling. Furthermore, recent works based on these methods have emerged: NVDiffrecmc~\cite{hasselgren2022nvdiffrecmc} employs Monte Carlo Rendering for shading, building upon NVDiffrec's geometry and material representation, while SAMURAI~\cite{boss2022-samurai} additionally estimates camera pose for inverse rendering. In our comparisons, we also consider NeRFactor~\cite{zhang2021nerfactor} and InvRender~\cite{zhang2022invrender}, which are implicit representation-based methods utilizing distinct approaches for geometry and illuminations. To assess the image quality of relighting and albedo images, we utilize three quantitative metrics: Peak Signal to Noise Ratio (PSNR), Structural Similarity Index Measure (SSIM), and Learned Perceptual Image Patch Similarity (LPIPS). For evaluating geometry on synthesis datasets with ground truth geometry, we report the Chamfer Distance (CD).

\input{tables/tab_quant-exp}

\subsection{Implementation Details}

\textbf{Geometry Module.} To encode the geometry, we utilize multiresolution hash encoding \cite{muller2022instant} with 16 levels and 2 features per level. The coarsest resolution is set at 16, and the hash table size is $2^{19}$. The original position, scaled by a factor of 2 and offset by $-1$, is included in the positional encoding. The multi-layer perceptron (MLP) responsible for learning positional features after hash encoding consists of one hidden layer with 64 neurons activated by the Rectified Linear Unit (ReLU) function. The output positional feature dimension is set to 13. For geometric weight initialization, we adopt the approach used in NeuS~\cite{wang2021neus}. In the output layer, no activation function is applied as the first element of its output represents the SDF value, which should not be restricted within a specific range.

\textbf{Radiance Module.} The outgoing radiance is modeled by a Multi-Layer Perceptron (MLP) consisting of 2 hidden layers, each containing 64 neurons. The inputs to the MLP are concatenated and include the viewing direction (represented by sphere harmonics up to the $4^{th}$ level), positional features, and the unit normal vector of the surface. Rectified Linear Unit (ReLU) activation is applied to the hidden layers. The final output of the MLP is a 3-channel RGB radiance, which is further activated using the Sigmoid function, constraining the values within the range of $[0, 1]$.

\textbf{Material Module.} The spatially varying material field is modeled using an MLP with two hidden layers, each consisting of 64 neurons activated by the Rectified Linear Unit (ReLU) function. The positional feature computed by the geometry module serves as the input to this MLP. The output of the MLP is a 6-channel material feature, which is further activated using the Sigmoid function. In this 6-channel feature, the first three channels represent the diffuse albedo ($\vk_d$), while the remaining three channels represent the specular and occlusion properties ($\vk_s = {o, r, m}$) for occlusion, roughness, and metallic, respectively.
To represent the illumination, we utilize a differentiable cubemap with a resolution of $512 \times 512 \times 6$. During training, the pixel values of the cubemap are clipped to be greater than 0, ensuring valid illumination representation. For visualization purposes, the cubemap is converted into a $512 \times 1024$ High Dynamic Range (HDR) image using the lat-long conversion method.

\begin{figure*}
    \centering
    \includegraphics[width=\linewidth]{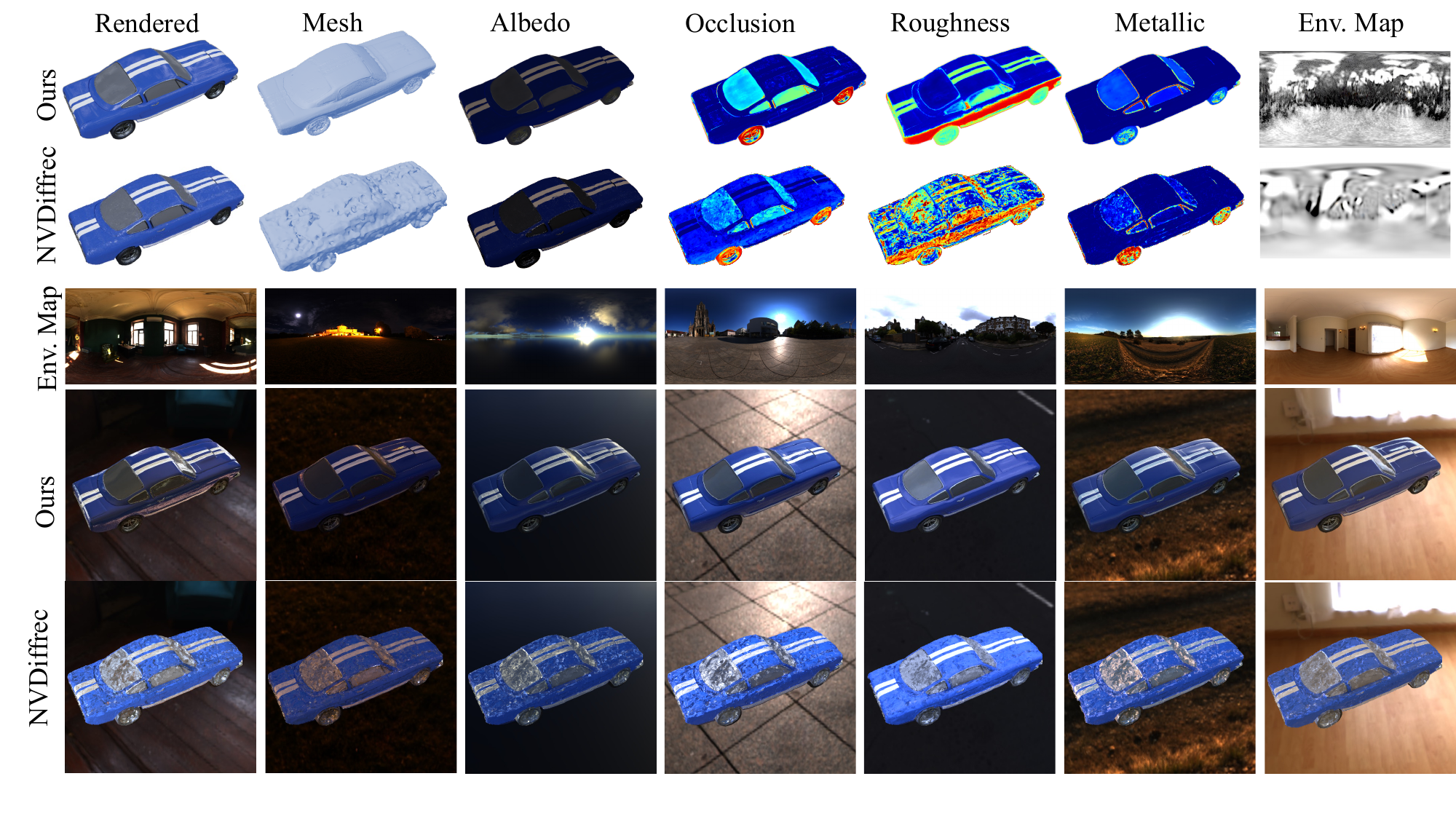}
    \caption{Material factorization and relighting on the shiny car scene in Ref-NeRF's Shiny dataset.}
    \label{fig:shiny_visualize_car}
\end{figure*}

\subsection{Training Details}
We perform joint optimization of scene geometry, material, and environmental illumination by incorporating both image loss and parameter regularization. For the image loss, we assign priority to Mean Squared Error (MSE) loss with specific weightings: $\lambda_{c1} = 1$, $\lambda_{c2} = 10$, and $\lambda_{mask} = 0.1$. Regarding parameter regularizations, we set $\lambda_{se}=0.1$, $\lambda_{l}=0.1$, $\lambda_{mio}=0.001$, and keep all other parameters at their default value of $0.01$. The optimization of our model is carried out using the Adam optimizer with a learning rate of $0.01$. The learning rate undergoes a scheduled progression, starting with a 500-step warming-up stage where it begins at 1\% and gradually increases to 100\%, followed by exponential decay until the end of training. Asymmetrical scheduling is applied to the material and radiance modules to facilitate geometry initialization. Our experiments were conducted on 2 NVIDIA Tesla V100 GPUs, with a training duration of approximately 1.5 hours, encompassing a total of 40,000 steps.

During training, we employ dynamic ray sampling to adjust the number of rays used for training. Initially, we set the number of rays for training to 256, with a maximum limit of 8,192. To ensure the cosine value between the viewing direction and the normal vector remains valid, we follow NeuS's convention of smoothing the cosine value using a cosine annealing ratio during the first 5,000 steps \cite{wang2021neus}.
For ray-marching, we utilize an occupancy grid with a resolution of 128 within a cube range of $[-\text{rad}, \text{rad}]^3$. This grid is updated at each step to remove empty spaces where the opaque density falls below a threshold of 0.001 \cite{li2022nerfacc}. To stabilize geometry learning, random values are assigned to masked areas during training.
During CO3D data training, all scenes are processed and recentered as described in \cite{insafutdinov2022snes}, and we use a bounding box with a radius of $1.2$. For training the Shiny and NeRFactor's relight dataset, we set the radius to $1.5$. The scaling factor $\tau$ for $\Phi_{\tau}(s)$ in calculating the opaque density is initialized as $e^3$ and updated using the Adam optimizer with a learning rate of 0.001.

\subsection{Experiment on Synthetic Dataset}

\textbf{NeRFactor's Relight Dataset.} Following NeRFactor \cite{zhang2021nerfactor}, four synthetic scenes originally released by NeRF \cite{mildenhall2020nerf} are relit with eight different low-frequency environment illuminations, and evaluated over eight uniformly sampled novel views. We compare our method with baseline NeRFactor~\cite{zhang2021nerfactor}, NVDiffrec \cite{munkberg2022extracting}, and NVDiffrecmc \cite{hasselgren2022nvdiffrecmc} on NeRFactor's Blender dataset on relighting and albedo reconstruction qualities. As material and fixed illumination can only be resolved up to a relative scale, we adopt the convention that scales the predicted albedo image by a color-tuning factor that matches the average of ground-truth albedo. {As shown in Tab.~\ref{tab:quant-exp}, our method outperforms both NeRFactor and NVdiffrec and achieves on-par relighting performance with NVDiffrecmc. We attribute the performance gain to adopting efficient implicit neural surface representation together with pre-integrated rendering, whereas NVDiffrecmc's performance gain is contributed by novel Monte Carlo shading.
We list NeuS here as a baseline of non-relightable methods for comparison. The qualitative results are visualized in Fig. \ref{fig:exp1-relight-viewgen}, where we render the image using a constructed environment map from a novel view and relight it with a given illumination.

\begin{figure}[t]
    \includegraphics[width=\linewidth]{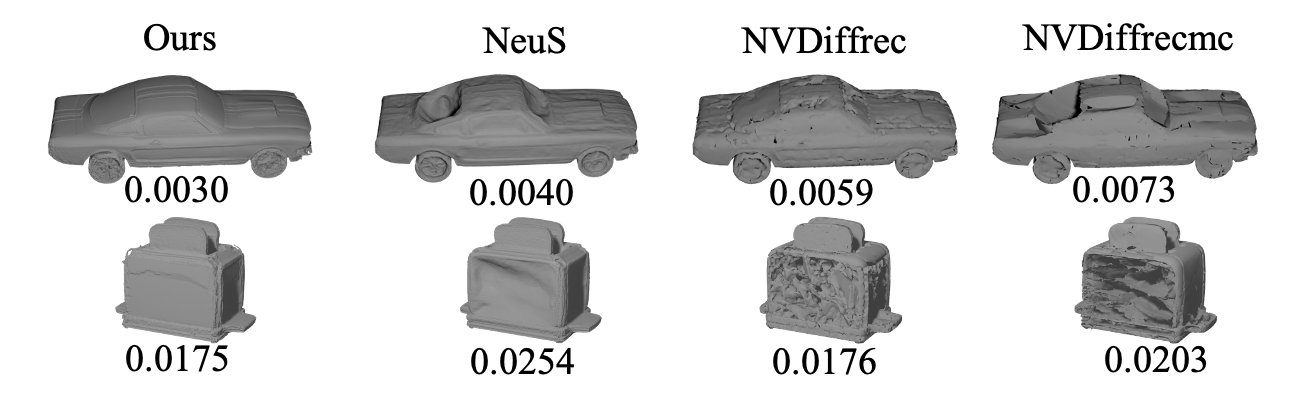}
    \caption{The evaluation of geometry on the Shiny Dataset and calculated Chamfer Distances for each example image. The Chamfer Distances are displayed below each image to indicate the quality of the geometry reconstruction.}
    \label{fig:quant-geometry}
\end{figure}

\textbf{Shiny Scenes.} 
As NeRFactor's Relight dataset contains mostly Lambertian surfaces illuminated by low-frequency environment light, we further evaluate our method on Ref-NeRF's Shiny scenes \cite{verbin2022ref}, which contains shiny objects. To evaluate the relighting ability in a more challenging scenario, We relight the Blender models of car and toaster using Blender Cycles renderer with 7 different high-frequency environment lights. 
Fig.~\ref{fig:shiny_visualize} shows the material factorization and relighting results on the toaster scene. The bread and toaster material were correctly predicted, and the relighting results blend in the novel illumination and exhibit consistent non-Lambertian reflection on the surface (notice the reflection of the environment map visible on the toaster).

We further show evaluations of the reconstructed geometry on the Shiny dataset in Fig.~\ref{fig:quant-geometry}. The Chamfer distance is calculated by uniformly sampling 5,000 points from the mesh and comparing them to the samplings from ground truth models.
Our method addresses the geometric ambiguity often found in NeuS by explicitly factoring the illumination, which makes it easier for the model to interpret shiny objects. Therefore effectively maintains the convex geometry on the highly reflective region of both the car and toaster model. Additionally, our method reproduces smoother surfaces compared to NVDiffrec/mc, showing the superiority of using continuous implicit neural surface representation over a discretized representation.

\subsection{Experiment on Real-World Dataset}

To evaluate our method on real-world scenes, we follow \cite{wang2021neus} in adopting Common Objects in 3D (CO3D) dataset~\cite{reizenstein2021co3d} and evaluating a subset of cars. The CO3D dataset is a collection of multi-view images captured in outdoor settings, containing detailed annotations such as ground-truth camera pose, intrinsic, depth map, object mask, and 3D point cloud. This dataset was gathered through real-world video capture and presents a significant challenge to reconstruction algorithms due to the presence of highly reflective and low-textured surfaces like dark windows, and metallic paint, which are non-Lambertian. As the ground-truth object mask is directly produced by off-the-shelf software, up to $8\%$ of the masks are wrong. In our experiment, we filtered out the incorrectly masked images by first computing the distribution of masked percentages of all images in a scene and then dropping the images whose masked percentage is below the second mode threshold if the distribution is multimodal. 

{We compare our method with NVDiffrec~\cite{munkberg2022extracting}, NVdiffrecmc~\cite{hasselgren2022nvdiffrecmc}, SAMURAI~\cite{boss2022-samurai}, NeuralPIL~\cite{boss2021neural}, and InvRender~\cite{zhang2022invrender} on a subset of 10 car scenes with relatively complete $360^\circ$ viewing directions in CO3D dataset. As shown in Tab.~\ref{tab:exp-co3d}, our results are significantly better than other methods of novel view synthesis. It is worth noting that NVDiffrecmc performs worse than NVDiffrec because NVDiffrecmc enforces additional regularization\cite{hasselgren2022nvdiffrecmc} and also tends to degrade the geometry as shown in Fig.~\ref{fig:quant-geometry}. A more detailed qualitative comparison with most related NVDiffrec methods is visualized in Fig.~\ref{fig:real-fact-relight}. Although its rendered novel views look realistic in the interpolated viewing direction, NVDiffrec fails to reconstruct a smooth mesh, and the geometry artifacts lead to noisy material factorization and unrealistic relighting results. We attribute this to the fact that NVDiffrec uses differentiable marching tetrahedrons with a fixed number of vertices to represent geometry, limiting its ability to represent geometry and behaving unstably under limited views. }

Figure~\ref{fig:co3d-more} presents qualitative results comparing our method with NVDiffrec and Neural-PIL. It is important to note that Neural-PIL utilizes a different material representation, so we do not visualize its roughness and specular maps using our jet colormap. In Neural-PIL, the roughness value ranges from 0 to 1 and is represented by black to white, while the specular value corresponds to the specular color.
Upon observation, we find that the environment map learned by Neural-PIL \cite{boss2021neural} lacks high-frequency details and tends to exhibit a blue-tinted bias. This could be attributed to the dataset used to train the environment illumination latent, which may have influenced this color bias. Our generated meshes, on the other hand, tend to preserve smoothness while capturing geometry details across most of the scenes. In contrast, NVDiffrec \cite{munkberg2022extracting} exhibits noise artifacts, while Neural-PIL \cite{boss2021neural} tends to exhibit an over-smoothed appearance.

\begin{figure*}
    \centering
    \includegraphics[width=\textwidth]{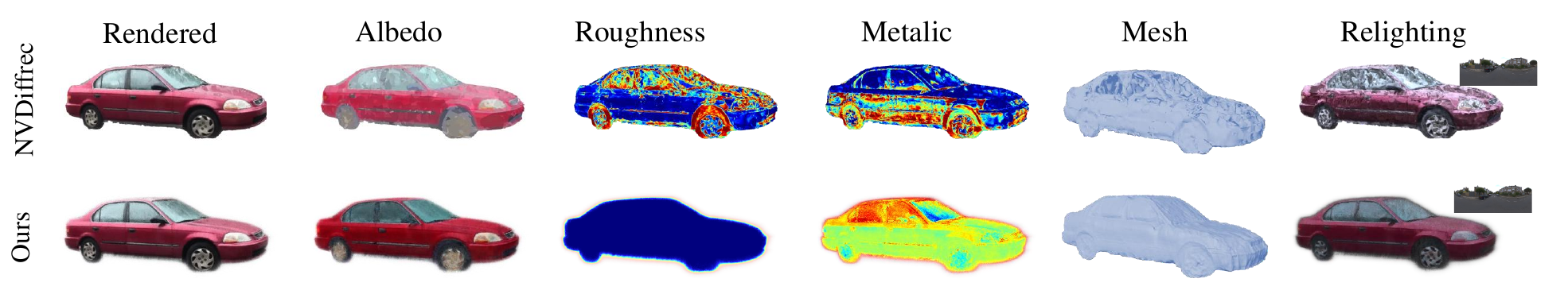}
    \caption{A comparison between our proposed method and NVDiffrec~\cite{munkberg2022extracting} on real-world CO3D dataset. Although NVDiffrec shows promising results on rendered novel views, it fails to reconstruct a plausible mesh. The geometry artifacts lead to noisy material factorization and unrealistic relighting.  our method with high-quality neural implicit field representation results in better overall results.}
    \label{fig:real-fact-relight}
\end{figure*}

\begin{figure*}
    \includegraphics[width=\linewidth]{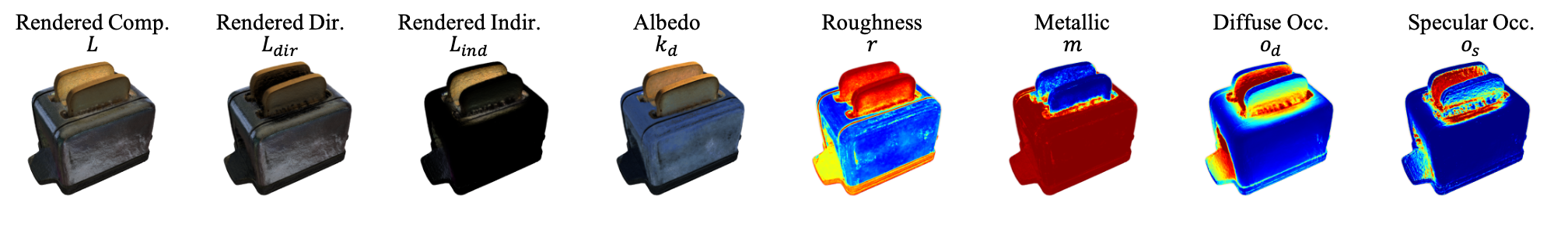}
    \caption{Indirect illumination distillation. The first three are rendered images using composed illumination $L$, direct illumination $L_{dir}$, and indirect illumination $L_{ind}$ respectively in Eq.~\ref{eq:indirect_compose}. The middle three images are fine-tuned material properties, and the last two images are distilled diffuse occlusion $o_d$ and specular occlusion $o_s$ for ~Eq.~\ref{eq:indirect_compose_dir}, and Eq.~\ref{eq:indirect_compose_indir} respectively. }
    \label{fig:indirect_illum}
\end{figure*}

\input{tables/tab_co3d}

\begin{figure}[t]
    \centering
    \includegraphics[width=\linewidth]{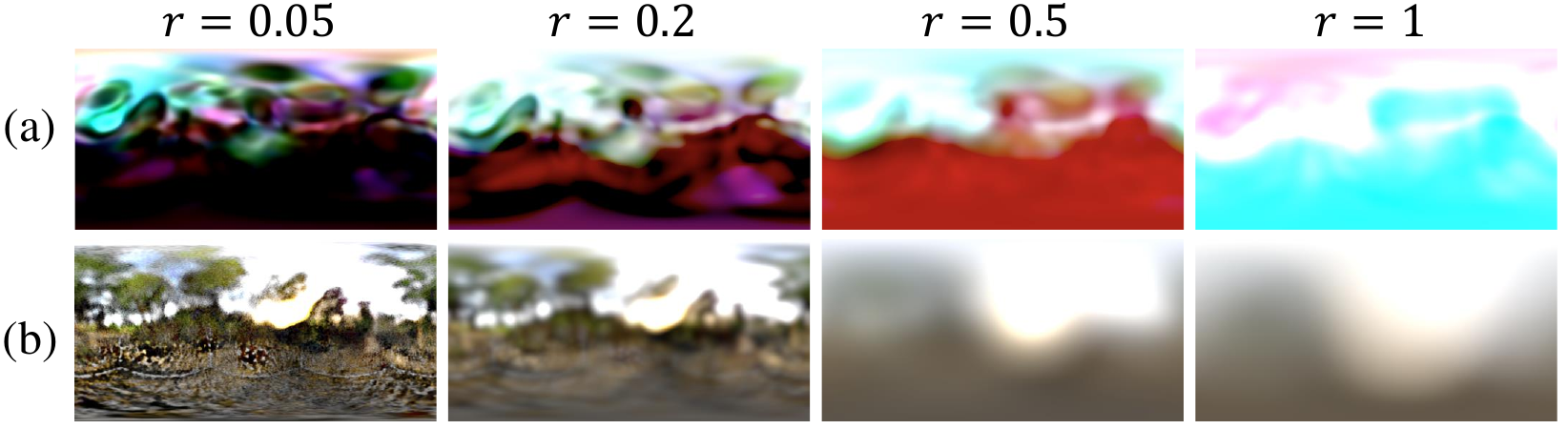}
    \caption{Different illumination modeling methods. (a) Using Neural-PIL~\cite{boss2021neural}'s environment illumination modeling. (b) Our proposed method.}
    \label{fig:light-modeling-compare}
\end{figure}

\subsection{Indirect Illumination}
We further show the additional indirect illumination distillation in Fig.~\ref{fig:indirect_illum} using the Toaster scene. The model is initialized using the pre-trained geometry, radiance and material module and fine-tuned for another 20K epochs by adding supervision for indirect illumination $\mathcal{L}_{ind}$ and occlusion $\mathcal{L}_{occ}$ as described in Sec.~\ref{sec:indirect_illum}. As shown in Fig.~\ref{fig:indirect_illum}, the learned occlusion terms for specular and diffuse occlusion help in distinguishing the contribution of direct and indirect illumination between the parallel breads as well as recognizing the indirect color bleeding on the shiny surface of the toaster machine. This decomposition leads to more consistent albedo prediction on the bread, eliminating the effect of complex indirect illumination. This shows that modular indirect illumination distillation can work effectively based on our learned representation.

\subsection{Ablation Study}
\label{ablation_study}

\textbf{Material and SDF Regularization.} We conduct an ablation study to compare the results on NeRFactor's dataset with and without a specific factor. The findings are summarized in Table~\ref{tab:quant-exp}. In terms of PSNR, we observe only marginal performance gains in relighting and albedo reconstruction. This can be attributed to the high-quality input views present in the dataset, which already provide regularization for material properties.
However, the inclusion of the SDF loss proves to be crucial in addressing the ambiguity between geometry and material properties. By penalizing noise in the reconstructed surface normals, the SDF loss plays a significant role. Consequently, the quality of the relighting results exhibits significant improvement when the SDF loss is employed.

\textbf{Neural-PIL vs. Pre-computed Environment map.}
We conduct a comparison between our approach of using a trainable environmental cubemap and explicitly computing pre-integration for lighting representation, and Neural-PIL that utilizes FILM-SIREN layers \cite{boss2021neural} to learn pre-integrated illumination with different roughness levels as network input. Our findings reveal that Neural-PIL does not enforce strict integration relationships among queried environment maps across various roughness levels.
To evaluate the effectiveness of Neural-PIL as an alternative illumination model, we assess its ability to reconstruct environment illumination using NeRF's material scenes. As depicted in Figure~\ref{fig:light-modeling-compare}, when queried at roughness 1, the environment map generated by Neural-PIL significantly differed from the pre-filtered environment maps at other levels and exhibited a complementary color. This erroneous prediction violates the inherent relationships expected across different roughness levels, consequently resulting in inferior material reproduction.
In contrast, our method learns high-quality illumination that remains consistent across different roughness levels, as specifically designed and demonstrated in Figure~\ref{fig:light-modeling-compare}. Furthermore, Neural-PIL relies on training an additional auto-encoder network to learn illumination latent codes using data-driven methods in order to regularize the illumination. In comparison, our method deduces illumination solely from the scene itself, eliminating the need for extra data and mitigating any induced bias, such as a blueish style caused by an overrepresentation of sky environment maps or lower resolution due to downsampling of the collected environment during auto-encoder training.

\input{tables/tab_sup-co3d-baseline.tex}

\textbf{Additional Regularization.} In Table~\ref{tab:sup-co3d-baseline}, we present an ablation study focusing on material and illumination regularizations for the CO3D dataset. The findings demonstrate a slight improvement in performance for novel view synthesis when employing reduced levels of regularization.

\textbf{Training Strategy.} To explore different training strategies, we incorporate the material module at various stages of training. In Table~\ref{tab:sup-co3d-baseline}, we present the results obtained by introducing the material module at 0, 1K, and 10K training steps. The findings indicate a slight improvement in performance when introducing the material module during the earlier phase of training. This can be attributed to the advantages of jointly optimizing geometry, material, and illumination in a single-stage approach.

\begin{figure*}
    \centering
    \includegraphics[width=\linewidth]{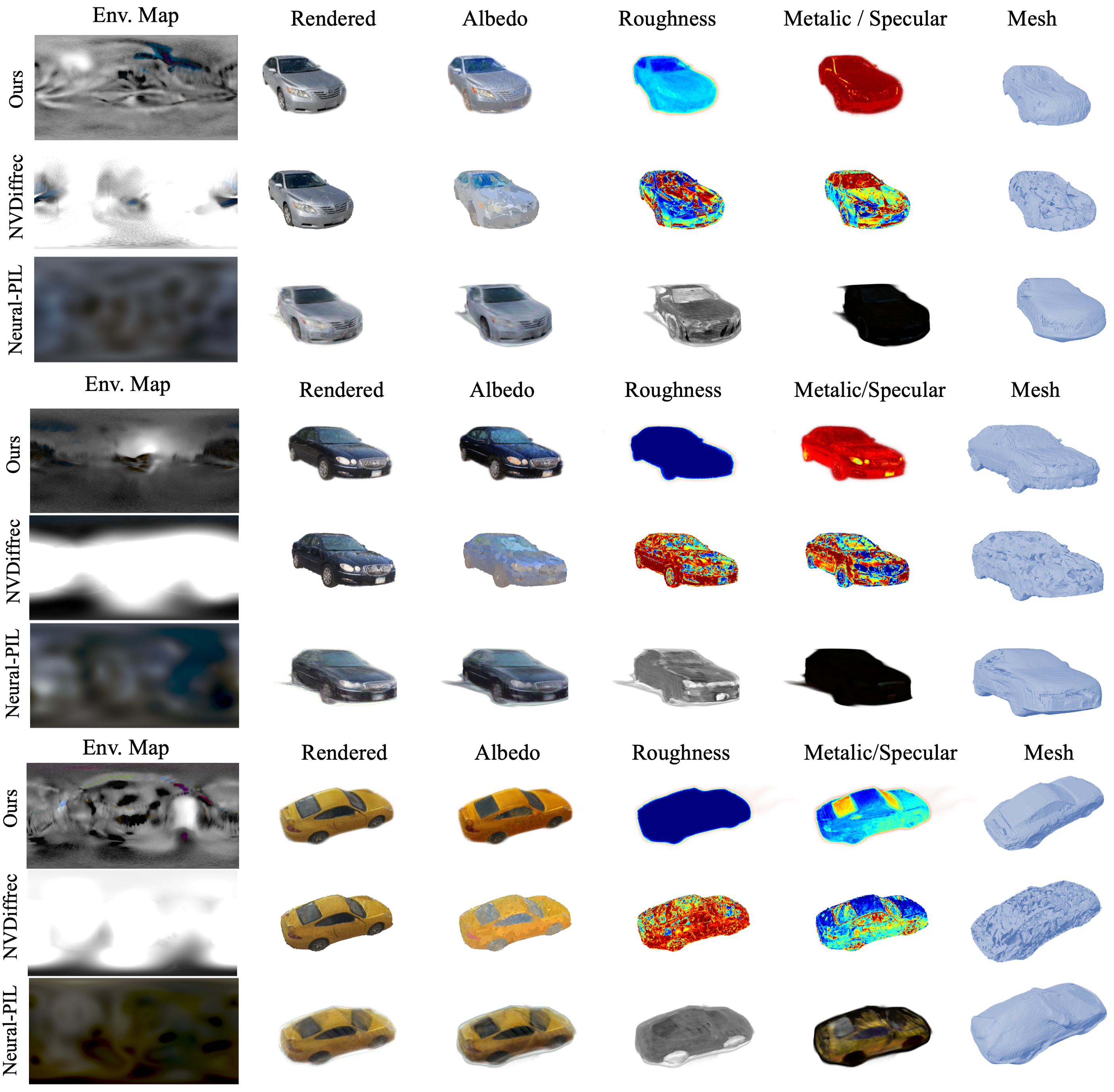}
    \caption{More quantitative results of scene car 351\_37072\_67647, 244\_25999\_52630, and  415\_57125\_110159 in CO3D dataset.}
    \label{fig:co3d-more}
\end{figure*}

\section{Conclusion and Discussion}
This paper presents a novel method called NeuS-PIR for effectively recovering relightable neural surfaces using pre-integrated rendering from images or monocular video. We propose simultaneously training a neural implicit surface, a spatially varying material field, and a differentiable environment cubemap using pre-integrated rendering. Additional indirect illumination fields can be distilled effectively from the learned representation. Our experiments demonstrate that our proposed method is superior to existing approaches in terms of reconstruction quality and relighting fidelity, which has the potential to be applied in many applications in computer vision and graphics.

\vspace{4pt} \noindent \textbf{Limitations.}~~We observe unsatisfactory quality when the input videos are short (\emph{i.e.}, the number of views is limited), which might be resolved by integrating data augmentations and diffusion models. While our method can handle wild data, it is sometimes not stable. Improving the robustness when dealing with wild data is also important. We use the marching cube algorithm~\cite{chernyaev1995marching} to export explicit representation (\emph{i.e.}, mesh), which may result in a loss of accuracy. Recent methods such as neural marching cube~\cite{chen2021neural} and neural dual contouring~\cite{chen2022neural} might help to export better explicit results for downstream applications. Additionally, current indirect illumination distillation uses geometry occlusion multiplication to approximate more complicated masked light integration.

\bibliographystyle{IEEEtran} 
\bibliography{tcsvt}

\vfill

\end{document}

%% file: tables/tab_quant-exp.tex
\begin{table}[!t]

\centering
\setlength{\tabcolsep}{1.2mm}
\renewcommand\arraystretch{1.1}

\caption{Quantitative evaluation on NeRFactor's synthesis dataset. Both NVDiffrec and NeRFactor metrics are as reported in~\cite{munkberg2022extracting} (In bold: best; Underline: second best). 
}  

\begin{tabular}{ l|c c c | c c c  }
\toprule
\multirow{2}*{Method}  & \multicolumn{3}{c}{Relighting} & \multicolumn{3}{|c}{Albedo}\\ 
{}& PSNR$\uparrow$ & SSIM$\uparrow$ & LPIPS$\downarrow$ & PSNR$\uparrow$ & SSIM$\uparrow$ & LPIPS$\downarrow$ \\
\toprule
 NeuS      & 21.83 & 0.913 & 0.070 & -     & -     & -      \\
 NeRFactor & 23.78 & 0.907 & 0.112 & 23.11 & 0.917 & 0.094 \\
 NVDiffrec & 24.53 & 0.914 & 0.085 & {24.75} & \underline{0.924} & 0.092 \\
 NVDiffrecmc &\underline{26.20} & \textbf{0.928} & \textbf{0.054} & \textbf{25.34} & \textbf{0.931} & 0.072  \\
 \hline
Ours       & \textbf{26.23} & \underline{0.925} & \underline{0.058} & \underline{24.86} & {0.921} & \underline{0.066} \\ 
w/o $\mathcal{L}_{mat}$  &26.07 & 0.923 & {0.062} & 24.60 & \underline{0.924} & \textbf{0.064}  \\
w/o $\mathcal{L}_{sdf}$  &24.05 & 0.904 & {0.085} & 24.61 & \textbf{0.929} & \textbf{0.065}  \\

\bottomrule
\end{tabular}\label{tab:quant-exp}
\end{table}

%% file: tables/tab_co3d.tex
\begin{table}[!t]
\centering
\setlength{\tabcolsep}{2.2mm}
\renewcommand\arraystretch{1.1}

\caption{Quantitative evaluation on CO3D dataset. ~\cite{reizenstein2021co3d}}  
\begin{tabular}{ l|c c c } 
\toprule
\multirow{2}*{Method}  & \multicolumn{3}{c}{Novel View } \\ 
{}& PSNR$\uparrow$ & SSIM $\uparrow$ & LPIPS $\downarrow$  \\
 \hline
   
 NVDiffrec~\cite{munkberg2022extracting}   & 26.29        & 0.925 & 0.086      \\
 NVDiffrecmc~\cite{hasselgren2022nvdiffrecmc}   & 24.45        & 0.911 & 0.107      \\
 SAMURAI~\cite{boss2022-samurai}   & 24.88        & 0.901 & 0.118      \\
 NeuralPIL~\cite{boss2021neural}   & 25.42        & 0.915 & 0.092      \\
 InvRender~\cite{zhang2022invrender}   & 24.94        & 0.919 & 0.092      \\
 Ours        & \textbf{29.03} & \textbf{0.935} &  \textbf{0.046}      \\
\bottomrule

\end{tabular}

\label{tab:exp-co3d}
\vspace{-10pt}
\end{table}

%% file: tables/tab_sup-co3d-baseline.tex
\begin{table}[!t]
\centering
\setlength{\tabcolsep}{3.6mm}
\renewcommand\arraystretch{1.2}

\caption{Ablation study on CO3D dataset~\cite{reizenstein2021co3d} }

\begin{tabular}{ l|c c c } 
\toprule
{Method}  &   PSNR$\uparrow$ & SSIM $\uparrow$ & LPIPS $\downarrow$  \\
 \hline

 NVDiffrec\cite{munkberg2022extracting}     & 26.29 & 0.925 & 0.086 \\
 Neural-PIL\cite{boss2021neural}    & 25.42     & 0.915  & 0.092     \\
 \hline
 Ours-0K       & 29.15 & 0.937 & 0.034      \\
 Ours-1K       & 29.13 & 0.938 & 0.033      \\
 Ours-10K (baseline)      & {29.03} & {0.935} & {0.046} \\
 \hline
 w/o $L_{mat}$   & 29.16 & 0.937 & 0.035      \\
 w/o $L_{light}$ & 29.10 & 0.937 & 0.034      \\
\bottomrule
\end{tabular}
\label{tab:sup-co3d-baseline}
\end{table}